\documentclass{article}
\usepackage{times}
\usepackage{hyperref}
\usepackage{url}
\usepackage{mathtools}
\usepackage{graphicx}
\usepackage{comment}
\usepackage{float}

\usepackage{amsthm}
\usepackage{amssymb}

\title{Sparse Neural Network Topologies}

\author{
Alfred Bourely \\
Columbia \\
\texttt{alfred.bourely@columbia.edu} \\
John Patrick Boueri \\
Columbia \\
\texttt{jpb2184@columbia.edu} \\
Krzysztof Choromonski \\
Columbia, Google Research \\
\texttt{kchoro@google.com} \\
}

\begin{document}

\maketitle

\begin{abstract}
We propose Sparse Neural Network architectures that are based on random or structured bipartite graph topologies.
Sparse architectures provide compression of the models learned and speed-ups of computations, they can also surpass their unstructured or fully connected counterparts. 
As we show, even more compact topologies of the so-called SNN (Sparse Neural Network) can be achieved with the use of structured graphs of connections between consecutive layers of neurons. 

In this paper, we investigate how the accuracy and training speed of the models depend on the topology and sparsity of the neural network.  
Previous approaches using sparcity are all based on fully connected neural network models and create sparcity during training phase, instead we explicitly define a sparse architectures of connections before the training. 
Building compact neural network models is coherent with empirical observations showing that there is much redundancy in learned neural network models. 
We show experimentally that the accuracy of the models learned with neural networks depends on "expander-like" properties of the underlying topologies such as the spectral gap and algebraic connectivity rather than the density of the graphs of connections.
\end{abstract}

\newpage
\section{Introduction}

The last decade has seen a rapid development of neural network methods that are widely considered to be the best tools for solving hard machine learning problems such as speech \cite{hinton},\cite{hinton2}, \cite{hinton3}, image \cite{lecun1}, \cite{krizhevsky}, \cite{ciresan} and video \cite{karpathy},\cite{simonyan} recognition. 
Neural networks are believed to learn  complex models by iterating two operations: linear projections of high-dimensional data and pointwise nonlinear mappings.

However, the process of building a good quality neural network models is time-consuming and the models' space complexity is usually very high. In this paper, we investigate how the quality of the neural network model depends on the topology of connections between consecutive layers.
We propose Sparse Neural Network (SNN) architectures that can match or exceed the accuracy of their dense counterparts. 
First, we study random sparse architectures that are quite compressible, then we study similar structured versions that provide even more compact description of the entire architecture. 
Sparse structured graphs are highly-compressible, yet they can share some properties with their sparse random counterparts.
We show good quality SNN model, with a well chosen topology, can equal or supersede its fully connected equivalent in accuracy, with a much smaller learned parameter space.
We also show that our results our robust as they hold over multiple layer neural network, or with the use of some over-fitting prevention techniques such as Dropout \cite{Srivastava}.
There is also additional insight into Neural Network inner workings and limitations by implementing SNN's and viewing them through the lens of graphs. We investigate how Algebraic Connectivity impacts a Neural Network's power, and show that there is a strong correlation between how interconnected a Neural Network is, and its resulting testing accuracy.
%%This makes it hard to port these models to devices with limited memory and computational resources such as mobile devices.%%In addition the advances in machine learning that were obtained with the use of neural networks,  the redundancy phenomenon that was empirically observed for neural network models persists. 
%%We have only recently started to understand how to compress these models in such a way such that the quality remains high, but computational speed-ups and space compression are still limited.

\section{Related work}

It was recently observed \cite{sindhwani}, \cite{felix}, \cite{kotagiri} that imposing specific structure of the linear embedding part of the neural network computations leads to significant speed-ups and memory usage reductions with only minimal loss of the quality of the model.
In this approach, the graph of connections between consecutive neural network layers is still dense and the reduction comes from recycling a compact vector of learned  weights across the entire matrix.
 
In \cite{novikov} the authors use a compact multilinear format, called the \textit{Tensor-Train format} to represent the dense weight matrix of the fully-connected layers. 
Even more basic technique relies on using the low-rank representation of the weight matrices. It was empirically tested that restricting the rank of the matrix of neurons connections to be of low rank does not affect much the quality of the model \cite{sindhwani2}, \cite{xue}, \cite{denil}, \cite{tai}. 
This setting is more restrictive than the one presented in \cite{sindhwani}, since low-rank matrices considered in these papers have in particular low displacement rank.

Other sets of techniques, such as methods exploting the so-called \textit{HashedNets} architectures, incorporate specific hashing tricks based on easily-computable hash functions to group weights connections into hash buckets.
This procedure significantly reduces the size of the entire model. 
Some other methods involve applying certain clustering and quantization techniques to fully-connected layers \cite{gong} that, as reported, have several advantages over existing matrix factorization methods.

Sparse neural network architectures were studied before, but in a different context. 
Good quality deep neural network/energy-based models might be obtained by imposing sparsity restrictions with nonzero activations \cite{glorot}, \cite{lee}, \cite{ranzato1} or using a linear rectifier as a nonlinear mapping.
Sparsity was also studied in the case of Dropout\cite{Srivastava} and DropConnect \cite{Wan} features where random vertices or edges are deactivated at each training batch to prevent over-fitting.

However, all of these architectures are based on a fully connected neural network that is sparsified during the training phase.
Our goal is not to deactivate many neurons or connection in the training phase by applying a particular regularization to the cost function, but rather to sparsify the graph of connections before. 
Our setting allows for speed up computations from the very beginning of the training phase due to the fact that sparse matrix - vector multiplication can be conducted much faster than its dense matrix - vector multiplication counterpart.

\section{Constructing sparse random and structured topologies}
We focus here on defining adjacency matrices of bipartite graphs between consecutive layers in the fixed neural network architecture.
Assuming that the $l^{th}$ layer consists of $n$ neurons and the $(l+1)^{th}$ layer consists of $m$ neurons, we consider the standard 
unweighted $n \times m$ adjacency matrix $\textit{Adj}_{l} \in \mathbb{R}^{n \times m}$, where:

\begin{equation}
Adj_{l}[i][j] =
\left\{
	\begin{array}{ll}
		1  & \mbox{if }  (i,j) \in E_{l}, \\
		0 & \mbox{otherwise,} 
	\end{array}
\right.
\end{equation}

$E_{l}$ denotes the set of edges of the corresponding bipartite graph and $(i,j)$ stands for the edge between $i^{th}$ neuron in the $l^{th}$ layer and
$j^{th}$ neuron in the $(l+1)^{th}$ layer.

Defined above adjacency  matrix is then applied element-wise to the weight matrix at the respective layer at each update, effectively zeroing out the non-desired 
connection.
Let $\delta_{l}$ denote the expected degree of the neuron in the $l^{th}$ layer, i.e. $\delta_{l} = \mathbb{E}[\sum_{j=1^{m}} \textit{Adj}[i][j]$.
To obtain satisfactory compression rates for our SNN, we set:  $\delta_{l}<m$.

Below we present specific structured topologies applied in our SNN as well as sparse unstructured topologies that were the subject of our analysis.
\subsection{Random Constructions}
\subsubsection{Random Edge Construction}
This is the most straight forward construction, where each node from the first layer is connected to independently chosen random set of nodes in the second layer, or to be more specific:
$$P[Adj[i][j]=1] = k/m$$
for some fixed parameter $k$. Notice that the number of edges of such a random bipartite graph has expectation $nk$ and highly concentrated around this expectation.
\subsubsection{Random Rotating Edge Construction}
In this construction, an edge matrix is created by a random binary vector of length $m$, and rotated at each row in a circulant fashion. The rotating vector is filled by assigning a connection with probability $$P[Adj[i][0]=1] = k/m$$
And we rotate this vector to get the connection matrix.
$$Adj[i][j'] = Adj[i][j]\ where\ j'=j-i\ mod\ m$$
To ensure the expected degree is in line with other constructions, we also fix the number of connections in the initial construction to be sure to have exactly $k$ connections. 
\subsubsection{Random d-Regular Expander Construction}
In this construction, an approximated bipartite expander graph is created. Each vertex from the input (or left layer) has a fixed degree $D = k/m$, meaning exactly $D$ connections with the output or right layer. The connections are then randomly assigned to the right layer, giving the random graph a high probability that it is an expander. For more details on probabilistic constant degree expanders refer to \cite{capalbo-02} 
\subsection{Structured Constructions}
\subsubsection{Regular Rotating Edge Construction}
In this construction, the deterministic vector of length $m$ is created by simply having $k$ ones followed by $m-k$ zeroes.
$$Adj[i][j]= 1\ for\ i\ mod\ m< j < i + k\ mod\ m,\ 0\ otherwise$$
This construction is particularly interesting due to its rigid structure, allowing for a better computing time, and a smaller memory space used. The edge matrix and its related graph is presented in the figure below.  
\begin{figure}[h]
	\begin{center}
		\fbox{\includegraphics[width=6cm,height=4cm]{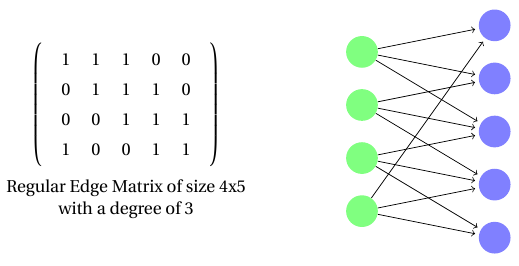}}
	\end{center}
	\caption{Experiments Results}
\end{figure}
\subsubsection{Long-Short Rotating Edge Construction}
In this construction, a deterministic vector of length $m$ is created by having $k/2$ ones followed by $k/2$ ones regularly placed between the $m - k/2$ positions left. The motivation being that connectedness of far away neurons allows for more information mixing and better accuracy than simple regular rotating edge construction. 
\subsubsection{Fibonnaci Rotating Edge Construction}
In this construction, a deterministic vector on length $m$ is created by using the $k$ first Fibonacci numbers. If the $k^{th}$ Fibonacci number is smaller than $m$, we create a connection if the index is in the Fibonacci list. We also fill the zero index with the 1st one of the Fibonacci list. If the $k^{th}$ Fibonacci number is bigger than m, we normalize our list by multiplying it with $m/Fibo(k)$. With this transformation, we get a list of float numbers. To be certain that we will have exactly $k$ ones in the vector, we fill the next element if the current value is already one. This vector is then rotated in a similar fashion as the previous deterministic constructions to produce the edge matrix.

\section{Experimental Results}
\subsection{Experimental Setup}
In order to investigate how sparsity impacts performance, we use the canonical MNIST database of 60,000 handwritten digits images as the training set, and an additional 10,000 images as the test set. The neural architecture we use is one hidden layer of size 100, 300, or 500, with the conventional flattened 28*28 input and 10 output layer size. We then train many networks with the varying constructions described above and varying degree k imposed on both the input to hidden layer connection and hidden layer to output, fitting a total of 3000 networks.
We train each network for 50 epochs with a fixed batch size of 32. We use mean square error as the loss function, the initialization function is a Glorot Gaussian \cite{glorot2010understanding}. A standard stochastic gradient descent optimizer is used with Nesterov momentum of 0.9 and  learning rate of 1\%. The activation function at all neurons is a regular sigmoid. 
Implementation was achieved using the python package Keras\cite{chollet2015} with Theano\cite{bergstra+al}\cite{Bastien-Theano-2012} as a backend. We then modify the update steps to impose sparsity at each step as described previously.
\subsection{Sparse Neural Nets Perform Well}
A key finding is that at very low levels of imposed sparsity, SNN can perform as well as their Fully Connected counterparts. In Figure 2, one can see that there is no precipitous drop off of accuracy until well below 10\% of the connections remain while varying the expected degree $k$ of the input layer to the hidden layer.  This result is consistent with varying the expected degree of the hidden layer as well, as seen in the contour plot of Figure 2. The experiments depicted are with a hidden layer of 300 and with a regular graph construction for simplicity, but the same result holds for all constructions and hidden layer sizes.
\begin{figure}[h!]
	\begin{center}
		\fbox{\includegraphics[width=12cm,height=6cm]{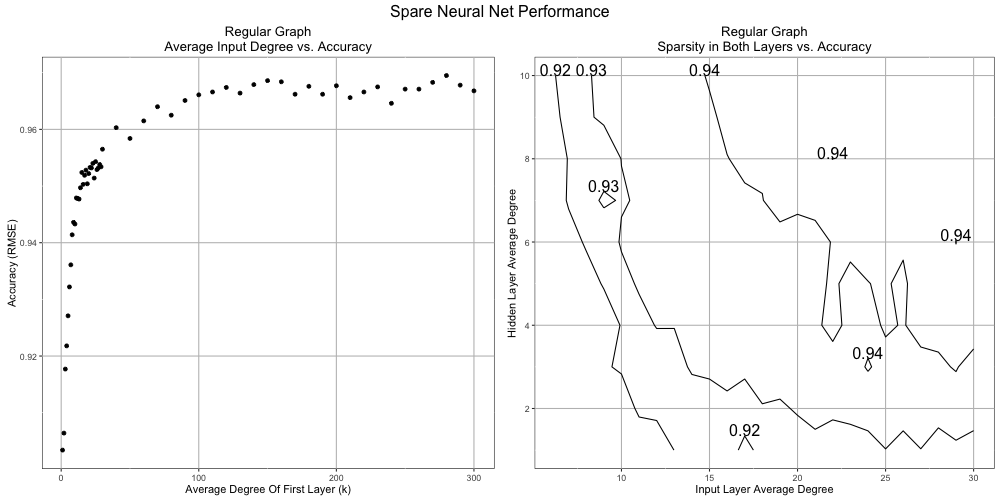}}
	\end{center}
	\caption{Experimental Results of a Regular Graph Construction with Hidden Layer Size 300}
\end{figure}
\subsection{Sparse Constructions Perform Differently}
Another key result is that different sparse architectures with the same degree of connection don't result in the same accuracy. In Figure 3, one can see how constructions vary in performance at very sparse levels on the left, as well as for increasing levels of connectedness on the right. In some cases, such as the structured Regular and Fibonacci constructions, the SNN outperforms its fully connected counterpart, indicating a better fitted model as a result of reducing the parameter space. The reason might be that big fully connected networks are worse at handling the noise because of their very redundant connections. Also, in a subsequent section we discuss how algebraic connectivity could be a possible way to predict this result.
\begin{figure}[h!]
	\begin{center}
		\fbox{\includegraphics[width=12cm,height=6cm]{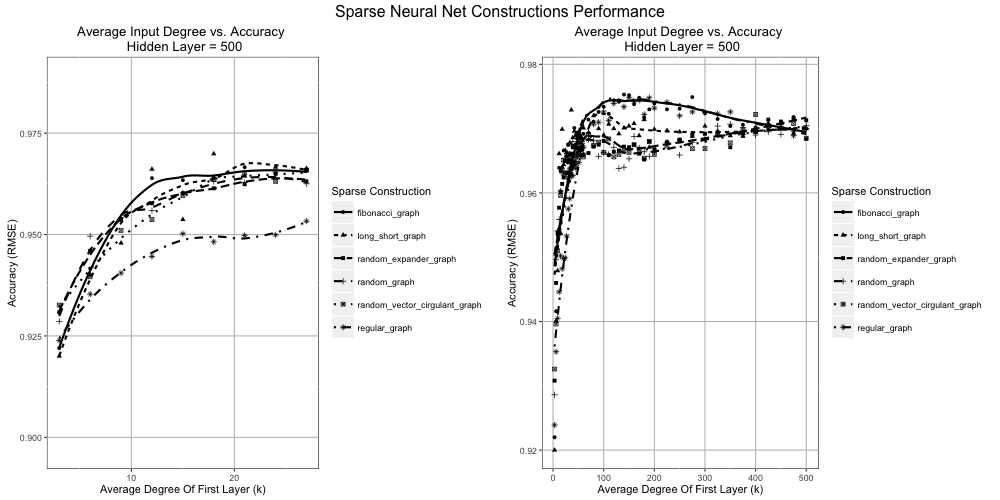}}
	\end{center}
	\caption{Varying Performance of Neural Net Construction}
\end{figure}
Overall, the Fibonacci SNN seems to be the best choice of structure. Indeed, it’s the only SNN that is both pseudo regular (which gives in good and sometimes better accuracy in the sparse setting) and pseudo random (which avoid sharp drop in accuracy if the network is very sparse). Regular SNN is a good second choice since it provides equivalently good results in most cases with a simpler implementation.
\subsection{The Results Hold With Dropout or Deeper Networks}
As previously discussed, over fitting is a big issue with neural networks and dropout methodology is extensively used to prevent this problem. We experimented with both sparsifying methods. We augmented our methodology of a SNN architecture which is set from the beginning with \cite{Srivastava}methodology of dropping a fixed number of vertices at each learning batch to avoid over fitting. With dropout=0.2 for the input layer and dropout=0.5 for the hidden layer, we show that the two methods are independent in Figure 4. The sparse Fibonacci and Regular networks still have a better accuracy than that of the fully connected equivalent. 

As neural networks rarely have only one hidden layer, and are becoming deeper with the increase in computing power we studied SNN with two hidden layers. In this setting, the structured Regular or Fibonacci SNN still performs better than its fully connected network equivalent. In Figure 4, we also present the results for a network of size 784-500-300-10 with the degrees of the two hidden layers varying at the same pace and the last layer being fully connected. In this case, the regular and Fibonacci sparse neural network still perform better than their fully connected equivalent. With only 30\% connections in the network, we are able to increase accuracy by 0.15\% after the same number of training epochs.

\begin{figure}[h!]
	\begin{center}
		\fbox{\includegraphics[width=12cm,height=6cm]{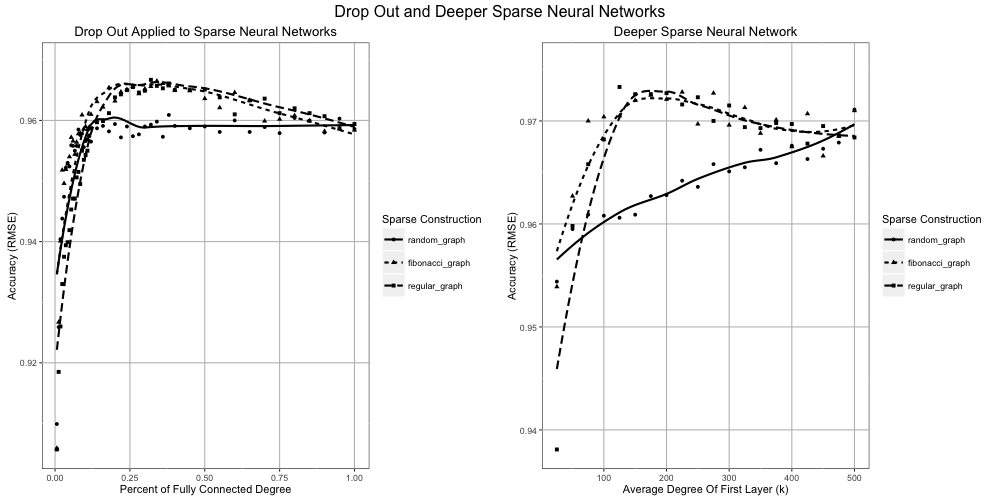}}
	\end{center}
	\caption{Sparse Constructions in Different Settings}
\end{figure}

\subsection{Graph Properties and Accuracy}
After training these SNN, there is a need to measure what causes the differences between topologies, and how graphical properties are related to the resulting accuracy. Two critical measures we identify are spectral gap and algebraic connectivity, which are defined as the first and second largest non-zero eigen value of the Laplacian matrix respectively. We find that connectivity varies between constructions substantially, particularly in the very sparse regimes, which can bee seen below. Hand in hand with this observation is connectivity is highly correlated with accuracy. When determining how to design a SNN topology, connectivity is a critical parameter to consider when constructing it.
\begin{figure}[h!]
	\begin{center}
		\fbox{\includegraphics[width=12cm,height=6cm]{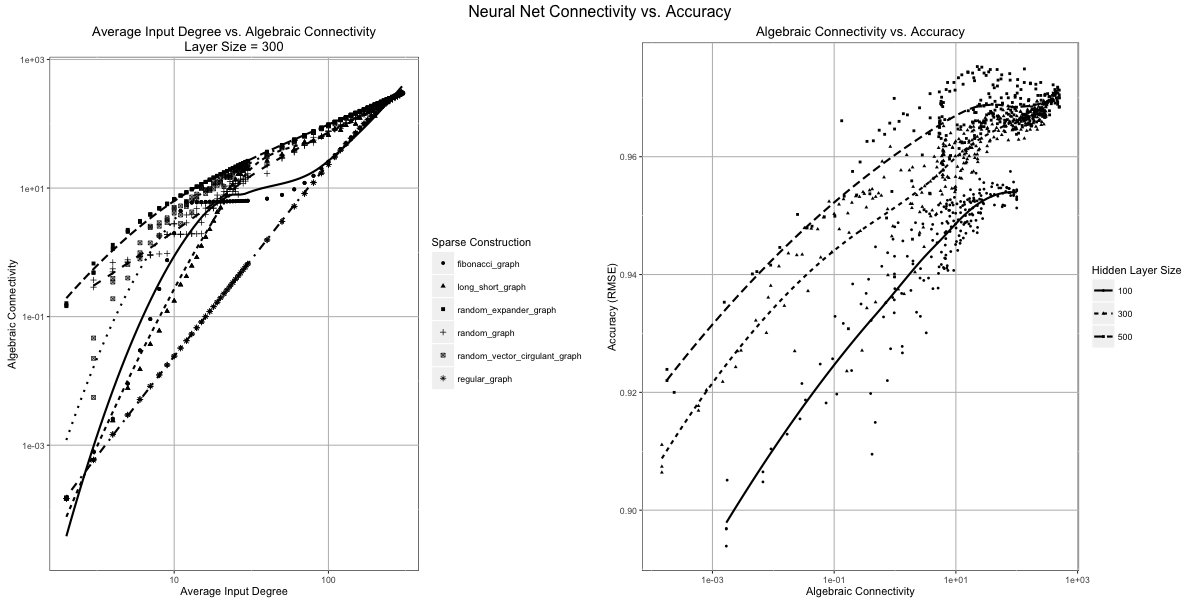}}
	\end{center}
	\caption{Connectivity of various constructions, and its relationship to accuracy}
\end{figure}
\subsection{Sparse Neural Network Learned Weights}
It is also interesting to look at how the learned weights vary with sparsity. While construction doesn’t have a particularly large impact on what the magnitude and distribution of the learned weights are, sparsity does. The more connections there are the more tempered the distributions become, finally leveling off at an equilibrium that reflects the same pattern that the overall accuracy of the SNN follows. The maximum, minimum and standard deviation of the connected weights lower with respect to both hidden layer size and degree connectivity. This follows intuition: the less weights you have, the more critical and high valued they become. 
\begin{figure}[h!]
	\begin{center}
		\fbox{\includegraphics[width=12cm,height=6cm]{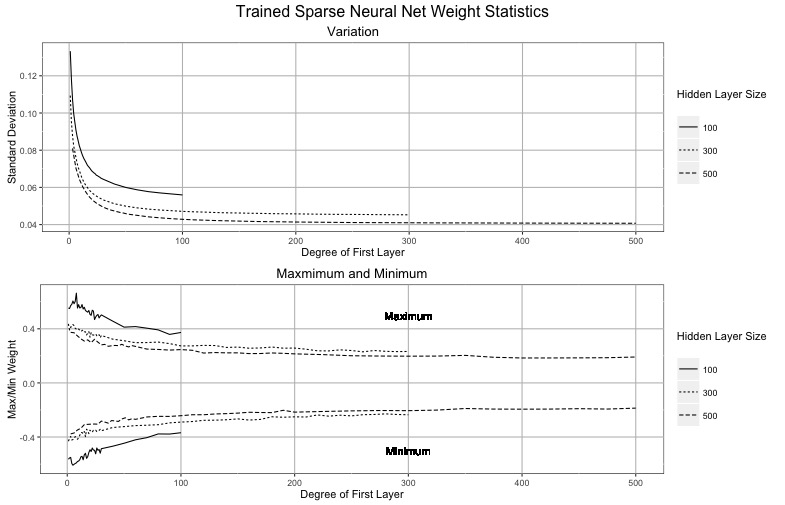}}
	\end{center}
	\caption{Trained Sparse Neural Net Weight Statistics}
\end{figure}

\newpage
\section{Conclusion and Future Work}
We have shown that SNN (Sparse Neural Network) constructions can be as competitive as, and sometimes outperform their fully connected counterparts with also shorter training time. The SNN can outperform its fully connected equivalent in the case of structured SNN such as a Regular or Fibonacci, with medium sparcity on large hidden layers. Overall, Fibonacci SNN different from Regular SNN as they are more accurate in very sparse setting (less than 10\% connections), but regular SNN also provide much simpler implementation.

We empirically verified that the accuracy of the sparse topologies depend on ”expander-like” properties of the matrix such as its algebraic connectivity. Much future work remains to further expand on the findings of this paper, such as applying the same techniques to other neural network architectures (i.e. as CNN, LSTM). This paper opens an orthogonal direction of study that promises interesting advances in neural network’s understanding and adds new tools to the neural network tool kit. It could eventually be employed to fit better models or provide tools for resources constrained applications such as connected and mobile devices.

\newpage
\small{
\bibliographystyle{unsrt}
\bibliography{sparse_neural_networks}
}
\end{document}